\documentclass[11pt, a4paper]{google}

\usepackage[authoryear, sort&compress, round]{natbib}
\usepackage{multirow}
\usepackage{xspace}
\usepackage{xurl}
\usepackage{tabularx}
\usepackage{graphicx}
\newcolumntype{C}{>{\centering\arraybackslash}X}
\usepackage[most]{tcolorbox}
\usepackage{listings}
\usepackage{float}
\lstdefinestyle{judgeprompt}{
  basicstyle=\ttfamily\footnotesize,
  breaklines=true,
  breakatwhitespace=true,
  columns=fullflexible,
  showstringspaces=false,
  backgroundcolor=\color{gray!8},
  frame=none,
  framexleftmargin=8pt,
  framexrightmargin=8pt,
  framextopmargin=4pt,
  framexbottommargin=4pt,
  xleftmargin=8pt,
  xrightmargin=8pt,
  aboveskip=8pt,
  belowskip=8pt,
}
\usepackage[capitalize, noabbrev]{cleveref}

\newtcolorbox{modelthought}{
  colback=gray!8,
  colframe=gray!8,
  boxrule=0pt,
  arc=2pt,
  left=8pt, right=8pt, top=6pt, bottom=6pt,
  fontupper=\ttfamily\small,
}

\paperurl{}

\logolist{logo_WHU, logo_NTU_EEE, logo_SEU, logo_CAS}

\title{SAMatcher: Dense Co-Visibility Modeling via Cross-View Fusion for Scale-Imbalance Image Matching}

\correspondingauthor{zhengxw@whu.edu.cn}
\reportnumber{} %

\author[1,2,\authfn{$\dagger$},\authfn{$\ddagger$}]{Xu Pan}
\author[1,\authfn{$\dagger$}]{Zhen Pang}
\author[1]{Qiyuan Ma}
\author[3]{Wei Ji}
\author[4]{Shuhan Shen}
\author[1,*]{Xianwei Zheng}

\affil[1]{State Key Laboratory of Information Engineering in Surveying, Mapping and Remote Sensing (LIESMARS), Wuhan University, Wuhan, P.R. China}
\affil[2]{School of Electrical and Electronic Engineering, Nanyang Technological University, Singapore}
\affil[3]{National Key Laboratory of Space Target Awareness, Space Engineering University, Beijing, P.R. China}
\affil[4]{Institute of Automation, Chinese Academy of Sciences, Beijing, P.R. China}

\authornote{$\dagger$}{These authors contributed equally to this work.}
\authornote{$\ddagger$}{Work done while at State Key Lab. LIESMARS, Wuhan University.}

\begin{abstract}
Reliable correspondence estimation supports image processing and 3D vision tasks, including Structure from Motion, visual localization, and image registration.
Wide-baseline matching is difficult under cross-view scale imbalance.
The same content may appear at different resolutions and spatial extents, creating unequal observation quality.
Such pairs constitute low-quality multi-view data, even when each image is clear.
Image information that is distinctive in one view may be weak or unavailable in the other.
Existing methods rely on pixel- or patch-level appearance and do not explicitly identify where reliable evidence is shared across views.
We propose SAMatcher, a modular co-visibility framework for scale-imbalanced image matching.
It uses symmetric cross-view interaction to fuse the two feature representations and identify their shared spatial support.
From the fused features, SAMatcher predicts dense co-visible masks and bounding boxes.
These predictions define view-specific cropping windows for downstream point-level matching.
Built upon the Segment Anything Model (SAM), SAMatcher extends monocular region modeling to cross-view co-visibility reasoning.
Point-sampled mask learning, box regression, and mask--box consistency jointly supervise its multi-granularity predictions.
Extensive experiments follow an evaluation protocol centered on cross-view scale imbalance.
The results show that SAMatcher consistently improves matching robustness across diverse matching pipelines.
When integrated with RoMa, SAMatcher improves AUC@5 and mAA@5 by 5.58 and 4.97 points, respectively.
The results demonstrate that cross-view fusion and dense co-visibility modeling provide reliable priors for matching under scale-induced quality imbalance.
Code and project page are available at \href{https://xupan.top/projects/samatcher}{xupan.top/projects/samatcher}.
\end{abstract}
 
\begin{document}

\maketitle

\section{Introduction}

\begin{figure}[htbp]
\centering
\includegraphics[width=0.8\columnwidth]{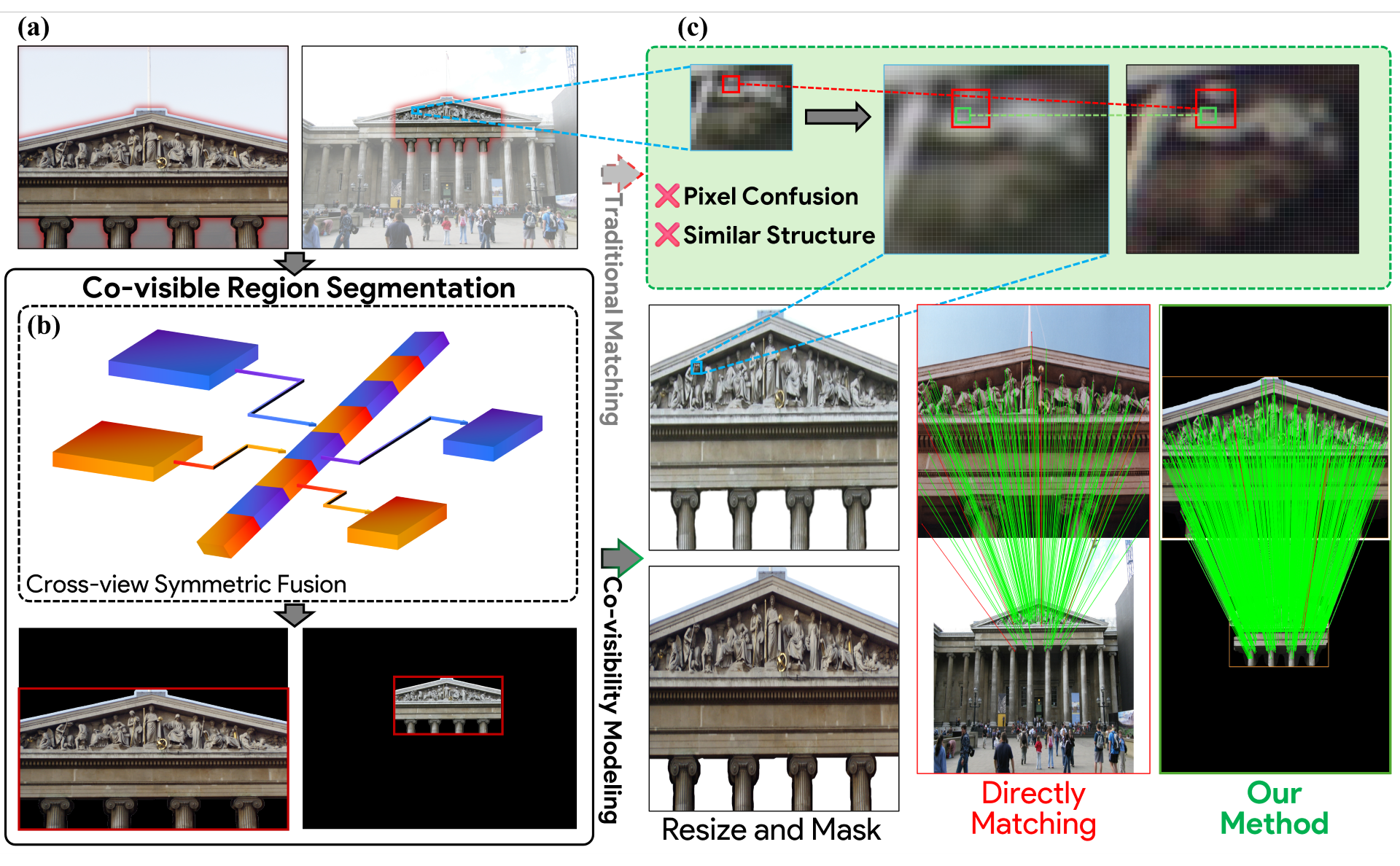}
\caption{
    \textbf{Motivation of SAMatcher under cross-view scale imbalance.}
    (a) Corresponding structures may occupy substantially different spatial extents across views, resulting in unequal observation quality in terms of effective resolution and spatial support.
    (b) This data-quality imbalance makes local regions with the same nominal size insufficiently comparable and increases correspondence ambiguity.
    (c) SAMatcher explicitly fuses cross-view features to identify regions whose visual evidence remains jointly reliable, providing co-visible cropping regions for downstream correspondence estimation.
}
\label{fig:intro}
\end{figure}

Robust feature matching is a fundamental problem in image processing, underpinning a wide range of applications such as image registration, Structure-from-Motion (SfM)~\citep{schonberger2016structure}, simultaneous localization and mapping (SLAM)~\citep{mur2015orb}, and 3D reconstruction.
At its core, the task aims to establish geometrically consistent correspondences across images for reliable spatial reasoning and scene understanding.
Despite decades of research, correspondence estimation remains challenging in the presence of low-quality multi-view data caused by large viewpoint changes, occlusions, illumination variations, and texture ambiguities. Such low-quality data often lead to sparse or unstable matches and error accumulation in downstream geometric pipelines~\citep{agarwal2011building}.

Among these factors, large inter-view scale variation is particularly consequential because it creates a data-quality imbalance between views. A structure represented by many pixels in one image may be compressed into a much smaller spatial extent in the other, so the two observations contain unequal amounts of usable image information and nominally corresponding patches cover inconsistent physical supports. We refer to this pair-dependent asymmetry in effective resolution, information availability, and spatial support as \emph{cross-view scale imbalance}. Unlike blur or noise that degrades an individual image, scale imbalance is relational: either image may be globally clear, yet the pair provides observations of unequal quality for the same content. Scale-imbalanced pairs therefore constitute low-quality multi-view data for correspondence estimation; the problem lies in the input observations and cannot be reduced to descriptor design.

Conventional correspondence pipelines typically rely on hand-crafted local feature extraction such as SIFT~\citep{lowe2004distinctive}.
Correspondences are then established via descriptor matching and subsequently verified using geometric constraints~\citep{hartley2003multiple}. While effective in controlled environments, these methods degrade significantly in textureless regions, repetitive structures, and long-baseline scenarios~\citep{heinly2012comparative,zhang2025comatcher}. 
Recent learning-based approaches improve robustness either by learning stronger feature representations or performing end-to-end correspondence estimation with deep networks~\citep{detone2018superpoint,sarlin2020superglue,sun2021loftr}. 
However, as illustrated in Fig.~\ref{fig:intro} (a,c), most existing methods still rely primarily on local appearance similarity and implicitly assume cross-view comparability at the pixel or patch level. 
This assumption becomes fragile under scale imbalance, where corresponding structures occupy substantially different spatial extents across views. 
As a result, scene information available in the higher-resolution observation may be weak or absent in its lower-resolution counterpart, while nominally corresponding patches may cover inconsistent physical supports. This imbalance in observation quality weakens cross-view comparability and allows visually plausible but geometrically inconsistent regions to produce spurious matches.

Pairwise cross-view matching is the most basic and central component of multi-view correspondence: multi-image associations, feature tracks, camera estimation, and 3D reconstruction are constructed by linking observations across image pairs~\citep{hartley2003multiple,schonberger2016structure}. Addressing data-quality imbalance at the two-view level therefore provides a fundamental building block for robust multi-view systems.

A key observation underlying this work is that scale imbalance cannot be characterized from either view alone: the quality of one observation depends on how the same scene content is recorded in the other view. Stronger descriptors alone cannot recover image information that is absent from the lower-resolution observation. Many correspondence failures therefore stem from the lack of explicit cross-view reasoning about where usable information is jointly available. The objective is not to reconstruct missing details, but to identify the shared spatial support on which reliable correspondence can still be established from the imbalanced cross-view observations.

To address this limitation, we explicitly fuse cross-view features to infer regions that remain jointly usable despite the scale-induced imbalance in observation quality. Rather than directly predicting point-wise matches, we estimate region-level masks that identify areas jointly visible across views and use them as structured priors to guide downstream matching. 
As illustrated in Fig.~\ref{fig:intro} (b), co-visible region segmentation suppresses irrelevant or non-overlapping content and localizes regions that are meaningfully comparable across views. 
By constraining correspondence estimation to these regions (Fig.~\ref{fig:intro} (c)), the formulation reduces ambiguous matches and improves robustness under large viewpoint and scale variations.

Building upon this insight, we propose \textbf{SAMatcher}, a framework that explicitly models co-visible regions for correspondence estimation under cross-view scale imbalance. Unlike end-to-end dense matchers, which fuse cross-view information primarily for direct correspondence prediction, SAMatcher uses fusion to reason about the data-quality imbalance between observations and isolates dense co-visible region inference as a standalone module. The predicted regions enable downstream matchers to focus on jointly visible content that remains usable in both views, thereby reducing interference from non-overlapping regions and poorly supported content in the lower-resolution observation.

We leverage the region modeling capability of the Segment Anything Model (SAM)~\citep{kirillov2023segment}, adapting it from object-centric segmentation to co-visible region reasoning for correspondence estimation.
Given an image pair, SAMatcher first predicts co-visible region masks and their spatial extents prior to matching.
High-level features from both views are processed by a symmetric cross-view interaction module, enabling bidirectional information exchange while preserving view-specific structures.
Conditioned on the fused representations, a set of shared learnable prompts queries the decoder to localize regions that are jointly visible across views.
A unified decoder then predicts co-visible masks and corresponding bounding boxes for both images. The predicted regions are converted into view-specific cropping windows that restrict downstream matching to jointly visible image content.

To support reliable co-visible region learning under large scale and viewpoint changes, we introduce a structured supervision scheme that jointly enforces pixel-level accuracy, region consistency, and spatial consistency between masks and bounding boxes.

In summary, our main contributions are as follows:
\begin{itemize}
\item
We introduce dense co-visibility modeling for scale-imbalanced image matching. It identifies pixel-level shared support that remains reliable across unequally resolved observations and suppresses interference from non-covisible regions.

\item
We propose SAMatcher, a foundation-model-based framework that adapts SAM to fuse scale-imbalanced cross-view observations through symmetric bidirectional interaction and learnable region queries, producing dense co-visible regions rather than processing each view independently.

\item
We develop a multi-granularity mask--box representation with joint supervision. Experiments under cross-view scale imbalance demonstrate consistent improvements across classical, learning-based, and dense matching pipelines.
\end{itemize}

\section{Related Works}
\label{sec:related}

\subsection{Feature Matching and Correspondence Learning}

Feature correspondence estimation is a fundamental problem in photogrammetry and underpins many applications, including SfM~\citep{schonberger2016structure}, SLAM~\citep{mur2015orb}, and 3D reconstruction. 
Classical pipelines rely on hand-crafted local descriptors such as SIFT~\citep{lowe2004distinctive}, combined with geometric verification~\citep{hartley2003multiple}. 
While effective in controlled, well-textured environments, these methods degrade significantly under large viewpoint changes, repetitive structures, or low-texture regions~\citep{heinly2012comparative, agarwal2011building}.

The rise of deep learning has shifted focus toward learned feature representations and end-to-end correspondence pipelines, as summarized in recent surveys of local and multimodal image matching~\citep{xu2024local,jiang2021multimodal}. 
Early works such as LIFT~\citep{yi2016lift} and SuperPoint~\citep{detone2018superpoint} jointly learn keypoint detection and description, while SuperGlue~\citep{sarlin2020superglue} further enhances robustness by aggregating contextual information through graph neural networks. 
More recent detector-free approaches, notably LoFTR~\citep{sun2021loftr}, establish semi-dense correspondences using dense features and transformer-based correlation. 
Subsequent methods, including MatchFormer~\citep{wang2022matchformer} and LightGlue~\citep{lindenberger2023lightglue}, improve both accuracy and efficiency via hierarchical attention and correlation-based designs.

Despite these advances, most existing approaches remain predominantly pixel- or patch-level, relying on local similarity metrics or global attention mechanisms. 
They typically assume co-visibility implicitly and lack explicit mechanisms to identify jointly visible regions across views. 
As a result, correspondence estimation remains vulnerable to occlusions, drastic viewpoint or scale changes, and background clutter. 
These limitations motivate region-level priors that explicitly model co-visible areas to guide robust correspondence estimation, which lies at the core of our proposed framework.

\subsection{Region-Level Representations and Co-Visibility Modeling}

Region-level representations provide structured semantic and spatial context that complements point-wise features. Multi-view representation learning further distinguishes information shared across views from content unique to each observation~\citep{zhang2024discovering}. Modern segmentation frameworks provide dense masks for region-level visual reasoning. Transformer-based architectures such as Mask2Former~\citep{cheng2022masked} combine global context modeling with fine-grained mask prediction, while the Segment Anything Model (SAM)~\citep{kirillov2023segment} enables class-agnostic, prompt-driven segmentation across diverse scenes. These advances make segmentation-based representations a useful bridge between semantic reasoning and geometric localization.

Most segmentation methods, however, operate on individual images and do not enforce consistency between views. Cross-view correspondence instead requires identifying content that is jointly visible while suppressing occlusions, background clutter, and viewpoint-dependent appearance changes~\citep{jin2021image,ni2024eto}. Transformer-based matchers~\citep{sun2021loftr,jiang2021cotr} and learning-based pose pipelines~\citep{sarlin2021back} improve robustness through long-range context and implicit geometric constraints, but they still reason mainly at the pixel or patch level and do not explicitly delineate co-visible regions.

Region-level guidance has also been explored for correspondence estimation. OETR~\citep{chen2022guide} constrains matching with predicted bounding boxes, but its coarse regions do not provide precise boundaries or fine-grained cross-view consistency. Classical pipelines handle co-visibility implicitly through reprojection overlap and bundle adjustment~\citep{agarwal2011building}, while multi-view stereo methods incorporate visibility reasoning for task-specific reconstruction~\citep{wang2022mvsnet}. Explicit co-visible region modeling instead supplies an interpretable spatial prior that constrains the correspondence search space, which directly motivates SAMatcher.

\subsection{Geometry-Aware Supervision in Vision Tasks}

Geometry-aware supervision has emerged as an effective strategy for integrating visual perception with three-dimensional reasoning by enforcing consistency across views, modalities, or reconstructed structures. 
In correspondence learning, techniques such as cycle consistency, depth-guided warping, and epipolar-aware attention have been shown to improve robustness under large viewpoint and scale variations~\citep{sun2021loftr}. 
These constraints encourage geometrically plausible matches and reduce ambiguities arising from local appearance similarity.

Beyond feature matching, geometry-aware supervision is widely adopted in segmentation and detection tasks. 
Multi-view and multi-modal fusion frameworks combine geometric cues with semantic segmentation to enhance robustness under occlusions and viewpoint changes~\citep{wu2024joint, liu2024mvg}. 
These studies demonstrate that enforcing region-level geometric consistency through mask alignment, cross-view agreement, and box--mask coherence can substantially improve prediction reliability.

Inspired by these findings, we design a mask-level geometry-aware supervision scheme that enforces consistency between predicted co-visible regions under cross-view scale imbalance.
By integrating region consistency, point-sampled mask supervision, and box--mask regularization, our approach provides reliable guidance for correspondence estimation and further distinguishes our framework from existing pixel-level matching methods.

\subsection{Low-Quality Multi-View Data}

Multi-view learning integrates complementary observations to improve downstream inference, but conventional formulations often assume that all views are complete and similarly reliable~\citep{zhao2017multiview}. Real-world data frequently violate this assumption through missing observations, noise, or substantial quality differences between views. Research in \emph{Information Fusion} has addressed incomplete multi-view data by inferring missing views and by assigning unequal weights to views during representation fusion~\citep{wang2024incomplete,li2024incomplete}. It has also improved robustness to low-quality or unreliable views through evidential reasoning and reliability-aware aggregation~\citep{yue2025dissonance,chen2025rsea}. These studies show that robust fusion should account for view-dependent data quality rather than allowing unreliable views to dominate learning.

Our setting concerns a distinct but related form of low-quality data. Neither image is necessarily missing or globally degraded; instead, cross-view scale imbalance makes the same scene content unequally informative across the two observations. SAMatcher addresses this relational quality imbalance at the correspondence level by identifying jointly usable co-visible regions, rather than imputing missing views or fusing final decisions.

\section{Methodology}

In this section, we present the proposed \textbf{SAMatcher} framework for robust feature matching between scale-imbalanced cross-view observations. We first formalize the problem, then describe the overall architecture, the cross-view fusion module, the shared prompt mechanism, the tokenized decoder with a high-quality (HQ) branch, and finally the training objectives.

\subsection{Preliminary}
\label{sec:preliminary}

\subsubsection{Problem Definition}

We study cross-view correspondence between two RGB images
$\mathbf{I}_0, \mathbf{I}_1 \in \mathbb{R}^{3 \times H \times W}$ of the same scene from different viewpoints.
Beyond point-wise matches, our goal is to identify co-visible regions, which are image areas simultaneously visible in both views and that support reliable correspondence under viewpoint, scale, or occlusion variations.

Given an image pair $(\mathbf{I}_0,\mathbf{I}_1)$, SAMatcher predicts,
for each view $i\in\{0,1\}$, a co-visible mask
$\hat{\mathbf{M}}_i\in[0,1]^{H\times W}$ and its corresponding bounding
box $\hat{\mathbf{B}}_i\in\mathbb{R}^4$. The mask captures the
fine-grained and potentially irregular shape of the co-visible region,
whereas the bounding box provides a compact description of its
location and spatial extent.

We denote the ground-truth co-visible support in view $i$ by
$\Omega_i$. Since $\Omega_0$ and $\Omega_1$ lie in different image
coordinate systems, their consistency is established through
cross-view correspondences rather than spatial intersection. These
region predictions reduce the subsequent matching space and suppress
ambiguous content outside mutually visible areas.

\subsubsection{Overview}

SAMatcher adopts a fusion-centered formulation in which co-visibility is inferred from the interaction of cross-view representations rather than from independently encoded image content. Under scale imbalance, the input views differ in observation quality because they provide unequal local resolution, information availability, and spatial support. Cross-view interaction allows co-visibility inference to account for this data-level imbalance. Multi-scale features are extracted by a shared encoder $\Phi$ and fused by a cross-view module $\Psi$, producing embeddings $(\mathbf{E}_0, \mathbf{E}_1)$ that encode joint semantic context. Task-specific decoders predict masks and boxes from the fused embeddings; although differing in query design, they are jointly optimized to enforce semantic-geometric consistency.

The overall pipeline is summarized as
\begin{equation}
(\mathbf{I}_0, \mathbf{I}_1)
\;\xrightarrow{\Phi}\;
\;\xrightarrow{\Psi}\;
(\hat{\mathbf{M}}_0, \hat{\mathbf{M}}_1, \hat{\mathbf{B}}_0, \hat{\mathbf{B}}_1),
\end{equation}
producing co-visible-aware predictions that form the foundation for robust cross-view correspondence. Subsequent sections detail each component and training objective.

\subsection{Proposed SAMatcher}
\label{sec:proposed}

SAMatcher is a SAM-based co-visibility modeling framework designed to improve correspondence estimation across wide-baseline image pairs. It leverages semantic representations encoded by SAM and fuses information across views to infer dense co-visible regions. The predicted masks and bounding boxes characterize shared visibility and provide structured priors for downstream point-level correspondence estimation.

Given a source--target image pair $(\mathbf{I}_0, \mathbf{I}_1)$, SAMatcher infers co-visible region masks and their spatial extents in both views,
\begin{equation}
    (\hat{\mathbf{M}}_0, \hat{\mathbf{M}}_1, \hat{\mathbf{B}}_0, \hat{\mathbf{B}}_1)
    = \mathcal{F}_\theta(\mathbf{I}_0, \mathbf{I}_1),
\end{equation}
which serve as structured semantic and geometric constraints for downstream correspondence estimation.

\begin{figure}[htbp]
\centering
\includegraphics[width=\textwidth]{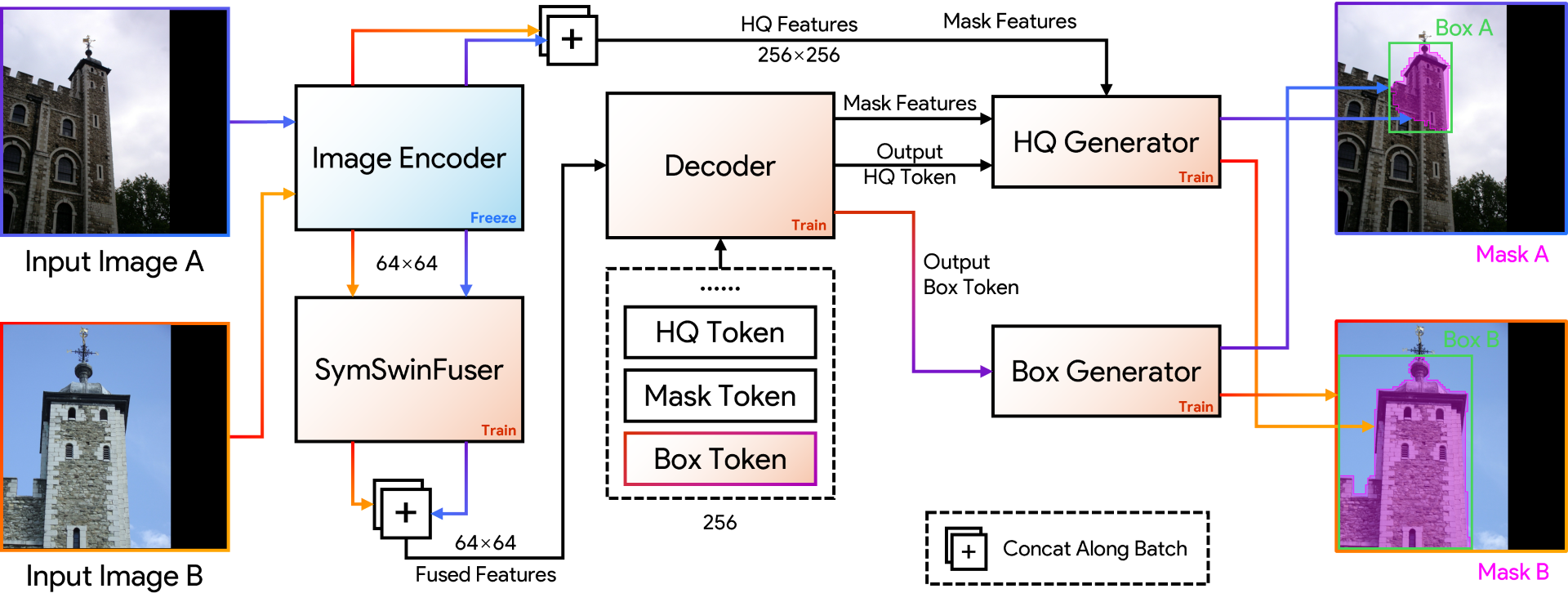}
\caption{
    \textbf{Overview of the proposed SAMatcher framework.}
    Given an image pair, SAMatcher extracts high-level visual representations using a shared encoder.
    A cross-view symmetric fusion module aligns semantic information across views and highlights potentially co-visible content.
    Based on the fused features, a prompt-driven mask decoder predicts co-visible region masks, while a dedicated box decoder estimates corresponding bounding boxes.
    These region-level predictions provide semantic and spatial priors that guide correspondence estimation, improving robustness under occlusion, background clutter, and large viewpoint or scale variations.
}
\label{fig:pipeline}
\end{figure}

As illustrated in Fig.~\ref{fig:pipeline}, both images are first processed by a shared visual encoder $\Phi$.
The architecture builds upon the high-quality variant of the Segment Anything Model (SAM-HQ)~\citep{ke2023segment, kirillov2023segment, ravi2024sam}, which provides strong mask decoding capacity and precise boundary modeling.
The encoder produces dense feature maps $\mathbf{F}_0 = \Phi(\mathbf{I}_0)$ and $\mathbf{F}_1 = \Phi(\mathbf{I}_1)$, capturing high-level semantic representations for each view.

To explicitly capture inter-view dependencies, the encoded features are processed by a cross-view symmetric fusion module $\Psi$, yielding fused representations
$(\tilde{\mathbf{F}}_0, \tilde{\mathbf{F}}_1) = \Psi(\mathbf{F}_0, \mathbf{F}_1)$.
Through bidirectional information exchange, this fusion step aligns semantic representations across views while preserving view-specific structures, and emphasizes regions that are likely to be jointly visible.

Unlike the original SAM formulation, which relies on explicit user-provided prompts, SAMatcher introduces a set of shared learnable prompts to implicitly condition region prediction.
This design is motivated by the observation that, in an image-pair setting, explicit prompts are neither available nor consistent across views.
Embedding prompting into learnable tokens enables image-pair-consistent self-prompting and cross-prompting behavior without external intervention.

Conditioned on the fused features and shared learnable prompts $\mathbf{P}$, co-visible region masks are predicted via the SAM mask decoder,
$\hat{\mathbf{M}}_i = \mathcal{D}_{\text{mask}}(\tilde{\mathbf{F}}_i, \mathbf{P})$, $i \in \{0, 1\}$.
These masks provide dense, pixel-level estimates of co-visible regions and constitute the primary semantic representation of co-visibility.

In parallel, spatial extents of the same co-visible regions are estimated by a dedicated box decoder operating on object-centric tokens.
Specifically, additional box tokens act as global queries and aggregate co-visible information via attention over the fused features.
Let $\mathbf{t}_{\text{box}}$ denote the learnable box tokens associated with each co-visible region.
Bounding boxes are decoded as
$\hat{\mathbf{B}}_i = \mathcal{D}_{\text{box}}(\mathbf{t}_{\text{box}};\,\tilde{\mathbf{F}}_i)$, $i \in \{0, 1\}$,
providing a compact geometric abstraction of the co-visible regions.
The parallel prediction of masks and boxes introduces complementary learning signals during training, encouraging consistency between semantic extent and geometric localization.

The predicted co-visible masks are converted into view-specific cropping windows and applied to the original input images. The resulting image pair is then passed to a pretrained downstream matcher. SAMatcher therefore learns only to extract co-visible regions, while correspondence estimation remains an independent downstream task. By restricting that task to jointly visible content, the predicted regions reduce interference from occluded and non-overlapping areas without changing the internal architecture of the selected matcher.

Overall, SAMatcher is a standalone, learnable co-visible region extractor. Cross-view fusion, prompt-driven mask prediction, and box localization are learned within SAMatcher, whereas point-level correspondences are produced independently by the chosen downstream matcher. This modular separation allows the same co-visibility model to provide semantic and spatial priors for different matching pipelines.

\subsection{Symmetric Cross-View Feature Interaction}
\label{sec:sym_swin_fuser}

Robust correspondence estimation under wide-baseline settings requires feature representations that simultaneously capture shared cross-view semantics and preserve view-specific structural cues.
Independent encoding of each view fails to model co-visibility, while overly aggressive fusion risks suppressing discriminative information critical under occlusion or partial overlap.
To address this trade-off, we propose a symmetric cross-view feature interaction module that explicitly models inter-view dependencies while maintaining representation fidelity for each view.

Recent advances in diffusion transformers and multimodal representation learning have shown that heterogeneous inputs can be effectively modeled as interacting token sequences within a unified attention framework \citep{peebles2023scalable}.
Inspired by this paradigm, we treat the source and target images as two strongly correlated modalities rather than independent observations.
This formulation enables token-level interaction across views, allowing semantic and geometric cues to be jointly contextualized without relying on explicit geometric parameterization.
Similar to multi-view representation frameworks that distinguish shared information from view-specific content~\citep{zhang2024discovering},
our formulation leverages attention-based interaction to implicitly align co-visible content across views.

Let $X_0, X_1 \in \mathbb{R}^{B \times L \times C}$ denote the flattened feature tokens extracted from the source and target images by a shared encoder, where $B$ is the batch size, $L$ the number of spatial tokens, and $C$ the feature dimension.
The goal of the interaction module is to produce cross-view-aware representations $\tilde{X}_0$ and $\tilde{X}_1$ that emphasize co-visible content while retaining view-specific details:
\begin{equation}
    (\tilde{X}_0, \tilde{X}_1) = \mathcal{F}_{\text{int}}(X_0, X_1).
\end{equation}

\begin{figure}[htbp]
\centering
\includegraphics[width=\linewidth]{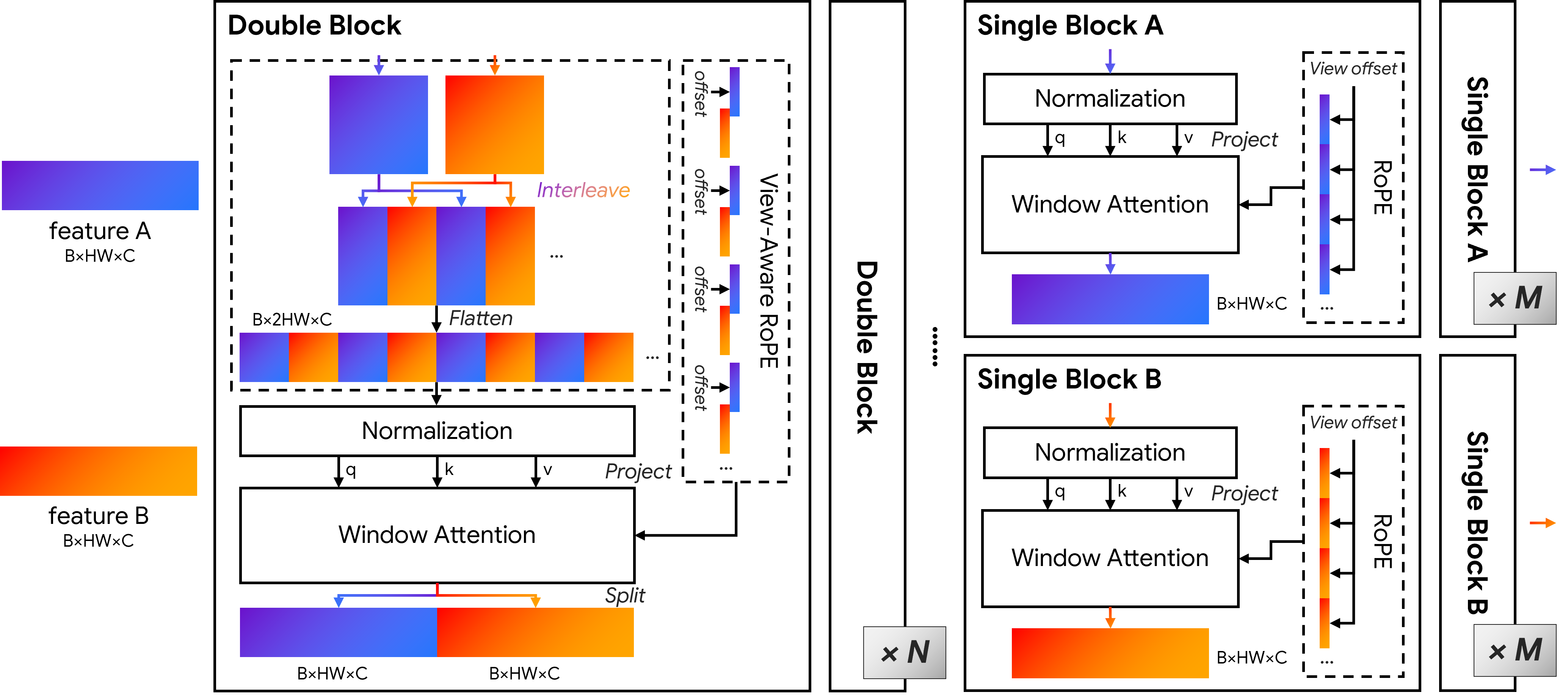}
\caption{
    \textbf{Architecture of the proposed symmetric cross-view feature interaction module.}
    Features from the source and target views are interleaved and processed by a stack of symmetric interaction blocks, enabling bidirectional token-level communication across views.
    Window-based attention with positional encoding facilitates efficient local interaction while preserving view identity.
    Subsequent single-view refinement further enhances view-specific structural details, producing cross-view aligned yet discriminative representations for downstream co-visible region prediction and correspondence estimation.
}
\label{fig:fuser}
\end{figure}

As illustrated in Fig.\ref{fig:fuser}, we first interleave tokens from the two views along the spatial dimension,
\begin{equation}
X_{\text{int}} = \operatorname{Interleave}(X_0, X_1) \in \mathbb{R}^{B \times 2L \times C},
\end{equation}
such that tokens from both views are jointly processed within the same attention windows. This interleaved sequence is processed by a stack of symmetric interaction blocks, organized into two stages, referred to as DoubleBlocks. Each stage consists of $L_d$ blocks.
Critically, to capture features at different levels of granularity, we progressively increase the number of attention heads across stages while maintaining a constant feature dimension.
The attention operator adopts windowed and shifted-window mechanisms from the Swin Transformer\citep{liu2022swin}.
For the interaction stages, we employ cosine attention to stabilize the training of cross-view correlations.

To preserve spatial coherence while enabling view discrimination, we incorporate view-aware rotary positional encoding into the attention computation.
Queries $Q$ and keys $K$ in the attention computation are rotated according to both spatial position and view identity,
\begin{equation}
    (Q', K') = \operatorname{RoPE}(Q, K \mid \text{view\_id}),
\end{equation}
allowing tokens from different views to remain distinguishable even when interleaved within the same attention window, where $\text{view\_id} \in \{0, 1\}$ indicates the source or target view.

Through multiple stages of DoubleBlocks, semantic information is progressively exchanged across views.
After the symmetric interaction stages, the interleaved sequence is split back into view-specific streams as $(X_0',X_1')=\operatorname{Split}(X_{\text{int}}')$.
These streams are subsequently refined by two additional stages of single-view attention blocks, referred to as SingleBlocks, with depths $L_{s1}$ and $L_{s2}$.
These blocks operate independently on each view using efficient Flash Attention~\citep{dao2022flashattention},
\begin{equation}
\tilde{X}_0 = \mathcal{R}(X_0'), \qquad \tilde{X}_1 = \mathcal{R}(X_1'),
\end{equation}
where $\mathcal{R}(\cdot)$ denotes the stack of single-view refinement stages.

By combining deep symmetric interaction with view-specific refinement, the resulting representations encode both shared co-visible semantics and discriminative local details.
These features serve as the input to downstream co-visible region prediction and correspondence estimation modules.
Importantly, the proposed interaction mechanism is conceptually independent of the downstream tasks and can be instantiated with different attention operators, highlighting its generality as a cross-view fusion strategy.

\textbf{View-Aware Rotary Positional Encoding.}
As illustrated in Fig.~\ref{fig:fuser}, positional encoding is an integral component of the proposed Symmetric Swin-based Cross-View Feature Fuser, as attention is performed over interleaved feature sequences originating from multiple camera views. In this setting, the positional encoding must simultaneously preserve spatial structure within each view and encode view identity to prevent ambiguity when features from different views are processed jointly. To satisfy these requirements, we adopt a \textit{view-aware Rotary Positional Encoding (RoPE)} tailored for interleaved cross-view attention.

Standard RoPE encodes relative positional information by applying position-dependent rotations to the query and key vectors. For an attention head with dimension $d_h$, the inverse frequency spectrum is defined as
\begin{equation}
    \text{inv\_freq}_k = \frac{1}{10000^{2k/d_h}}, \quad k = 0, \dots, \frac{d_h}{2}-1.
\end{equation}
Given a token at sequence position $p$, the corresponding rotation angles are
\begin{equation}
    \theta_p = p \cdot \text{inv\_freq} \in \mathbb{R}^{d_h/2}.
\end{equation}
Applying these rotations to the query and key vectors enables attention scores to depend on relative offsets, a property that has proven effective in transformer-based architectures.

However, in the proposed fuser, features from the source and target views are interleaved into a single sequence and processed within shared attention windows. In such a scenario, tokens from different views may occupy similar or even adjacent sequence positions, making standard RoPE insufficient to distinguish their origins. To address this issue, we introduce a small \textit{view-dependent offset} into the rotation angles. Specifically, for a token at position $p$ from view $v$, the view-aware rotation is defined as
\begin{equation}
    \theta_p^{(v)} = \theta_p + v \cdot \delta,
\end{equation}
where $v \in \{0,1\}$ denotes the source or target view, and $\delta$ is a fixed or learnable offset vector with the same dimensionality as $\theta_p$. This formulation preserves relative spatial encoding within each view while introducing a consistent phase shift that encodes view identity.

Following the standard RoPE formulation, each query vector $q \in \mathbb{R}^{d_h}$ is split into even and odd components, $q = [q_{\text{even}}, q_{\text{odd}}]$, and rotated as
\begin{equation}
    \begin{bmatrix}
        q_{\text{even}}' \\
        q_{\text{odd}}'
    \end{bmatrix}
    =
    \begin{bmatrix}
        \cos \theta_p^{(v)} & -\sin \theta_p^{(v)} \\
        \sin \theta_p^{(v)} & \cos \theta_p^{(v)}
    \end{bmatrix}
    \begin{bmatrix}
        q_{\text{even}} \\
        q_{\text{odd}}
    \end{bmatrix}.
\end{equation}
An analogous rotation is applied to the key vectors, while the value vectors remain unchanged. We denote this operation compactly as
\begin{equation}
    Q', K' = \text{ApplyViewAwareRoPE}(Q, K; v),
\end{equation}
which is used consistently in both the interleaved DoubleBlocks and the subsequent single-view refinement blocks.

By integrating view-aware RoPE into the attention computation, the SymSwin-Fuser can jointly model spatial relationships and view identity within a unified transformer framework. As visualized in Fig.~\ref{fig:fuser}, this design enables stable and discriminative cross-view interactions, ensuring that interleaved attention captures co-visible regions across views while maintaining view-specific spatial coherence under large viewpoint changes and partial visibility.

\subsection{Co-Visible Region Modeling and Decoding}
\label{sec:co_visible_decoding}

Given the cross-view fused features produced by the SymSwin-Fuser, SAMatcher performs co-visible region decoding through a heterogeneous yet tightly coupled decoding head.
The key design principle is to represent co-visibility at complementary granularities, combining dense spatial masks with object-level bounding boxes.
These two forms of region representation are decoded in parallel from a shared fused feature space, enabling consistent cross-view reasoning and mutual regularization during learning.

Let $\tilde{\mathbf{F}}_0, \tilde{\mathbf{F}}_1 \in \mathbb{R}^{B \times L \times C}$ denote the fused feature sequences for the source and target views, respectively, where $L = H'W'$ corresponds to flattened spatial locations.
As illustrated in Fig.~\ref{fig:pipeline}, SAMatcher adopts a prompt- and token-driven decoding strategy inspired by SAM-HQ, while extending it to the cross-view setting through explicit co-visibility modeling.

\vspace{0.5em}
\noindent\textbf{Spatial Queries for Dense Co-Visible Mask Decoding.}
To predict dense co-visible masks, the fused features are first projected into a shared query space:
\begin{equation}
    \mathbf{Q}_i = \tilde{\mathbf{F}}_i \mathbf{W}_q + \mathbf{b}_q, \quad i \in \{0, 1\},
\end{equation}
where $\mathbf{b}_q \in \mathbb{R}^{d_q}$ is a learnable bias shared across spatial locations.
Each query $\mathbf{q}_i^p \in \mathbf{Q}_i$ corresponds to a spatial location $p$ and encodes local appearance enriched by cross-view contextual information introduced by the symmetric fusion module.

Conditioned on shared learnable prompts $\mathbf{P}$, the mask decoder $\mathcal{D}_{\text{mask}}(\cdot)$ produces dense co-visible mask logits:
\begin{equation}
    \hat{\mathbf{M}}_i = \mathcal{D}_{\text{mask}}(\mathbf{Q}_i, \mathbf{P}), \quad
    \hat{\mathbf{M}}_i \in \mathbb{R}^{B \times L},
\end{equation}
which are reshaped and activated to obtain pixel-wise co-visible mask probabilities.
By operating on spatially grounded queries, the mask decoder preserves fine-grained boundary information while remaining explicitly conditioned on cross-view semantics.

\vspace{0.5em}
\noindent\textbf{Learnable Box Token for Object-Level Co-Visible Localization.}
While dense masks capture fine-grained co-visible regions, object-level localization requires a global representation not tied to specific spatial positions.
To this end, SAMatcher introduces a learnable box token $\mathbf{t}_{\text{box}} \in \mathbb{R}^{d_q}$, which is shared across views and concatenated with the prompt set $\mathbf{P}' = \{\mathbf{P}, \mathbf{t}_{\text{box}}\}$.
The box token serves as a global query that implicitly aggregates co-visible information through attention, rather than being directly derived from spatial features.

Specifically, the box decoder $\mathcal{D}_{\text{box}}(\cdot)$ treats $\mathbf{t}_{\text{box}}$ as query, attending to the fused feature sequence as key and value:
\begin{equation}
    \hat{\mathbf{b}}_i = \mathcal{D}_{\text{box}}(\mathbf{t}_{\text{box}};\, \tilde{\mathbf{F}}_i),
    \quad \hat{\mathbf{b}}_i \in \mathbb{R}^{B \times 4},
\end{equation}
where $\hat{\mathbf{b}}_i$ denotes normalized box coordinates in the corresponding view.
This formulation allows the box token to learn an object-centric representation by selectively attending to co-visible regions encoded in the fused features, without being constrained to any specific spatial location.

\vspace{0.5em}
\noindent\textbf{Parallel Decoding and Implicit Coupling.}
The dense mask decoder and the box token decoder operate in parallel on the same cross-view fused representations.
Although the box token is decoded independently from spatial queries, both prediction heads are implicitly coupled through shared prompts and joint optimization.
Dense masks provide fine-grained supervision over co-visible regions, while box predictions impose global spatial constraints on object extent.

This parallel decoding strategy enables complementary learning signals, encouraging consistency between pixel-level visibility and object-level localization.
As a result, SAMatcher produces multi-scale co-visible region representations,
\begin{equation}
    (\hat{\mathbf{M}}_0, \hat{\mathbf{M}}_1, \hat{\mathbf{B}}_0, \hat{\mathbf{B}}_1)
    = \mathcal{F}_{\text{decode}}(\tilde{\mathbf{F}}_0, \tilde{\mathbf{F}}_1),
\end{equation}
which jointly serve as structured semantic and spatial priors for robust correspondence estimation and mask--box consistency supervision.

\subsection{Joint Mask--Box Optimization and Supervision}
\label{sec:losses}

The SAMatcher head outputs co-visible masks and bounding boxes for each view, denoted as
\begin{gather}
\hat{\mathbf{M}}_i \in [0,1]^{B \times H_p \times W_p},
\qquad i=s,t, \\
\hat{\mathbf{B}}_i \in \mathbb{R}^{B \times 4},
\qquad i=s,t.
\end{gather}
where $B$ is the batch size, $H_p \times W_p$ is the predicted mask resolution, and bounding boxes are represented as $[x_\mathrm{min}, y_\mathrm{min}, x_\mathrm{max}, y_\mathrm{max}]$.
Ground-truth masks and boxes are denoted by $\mathbf{M}_i$ and $\mathbf{B}_i$, respectively.

To supervise mask prediction, we employ a point-sampled mask loss~\citep{kirillov2020pointrend} emphasizing uncertain regions.
For each view, let $\mathcal{P}_i \subset H_p \times W_p$ denote the sampled points based on predicted logits' uncertainty.
The point-sampled mask loss combines binary cross-entropy and Dice loss:
\begin{align}
    \mathcal{L}_{\mathrm{point}} &= \frac{\sum_{i \in \{0, 1\}} \sum_{p \in \mathcal{P}_i} \Bigl( \mathrm{BCE} + \mathrm{Dice} \Bigr) (\hat{\mathbf{M}}_i^p, \mathbf{M}_i^p)}{|\mathcal{P}_0| + |\mathcal{P}_1|}.
\end{align}

To incorporate object-level spatial structure, we further introduce a region-aware mask loss that assigns greater weight to pixels inside the predicted or ground-truth bounding boxes,
\begin{align}
    \mathbf{W}_i &= \mathbf{1}_{\mathrm{in\_box}} \cdot w_\mathrm{in} + (1-\mathbf{1}_{\mathrm{in\_box}}) \cdot w_\mathrm{out}, \\
    \mathcal{L}_{\mathrm{region}} &= \frac{\sum_{i \in \{0, 1\}} \sum_{p} \mathbf{W}_i^p \Bigl( \mathrm{BCE} + \mathrm{Dice} \Bigr) \bigl(\sigma(\hat{\mathbf{M}}_i^p), \mathbf{M}_i^p\bigr)}{\sum_{i \in \{0, 1\}} \sum_{p} \mathbf{W}_i^p},
\end{align}
where $\sigma$ denotes the sigmoid function, and $w_\mathrm{in} > w_\mathrm{out}$ prioritizes in-box pixels.

Bounding box supervision is provided by a dedicated box decoder, operating on learnable box tokens.
The box loss includes center and size regression, IoU-based alignment, and a mask--box consistency term that encourages the spatial extent predicted from masks to align with box predictions:

\begin{equation}
\mathcal{L}_{\mathrm{loc}}
=
\frac{1}{2B}
\sum_{i \in \{0,1\}}
\mathrm{L1}\!\left(
\mathrm{center}(\hat{\mathbf{B}}_i),
\mathrm{center}(\mathbf{B}_i)
\right),
\end{equation}

\begin{equation}
\mathcal{L}_{\mathrm{size}}
=
\frac{1}{2B}
\sum_{i \in \{0,1\}}
\mathrm{L1}\!\left(
\mathrm{size}(\hat{\mathbf{B}}_i),
\mathrm{size}(\mathbf{B}_i)
\right),
\end{equation}

\begin{equation}
\mathcal{L}_{\mathrm{IoU}}
=
\frac{1}{2B}
\sum_{i \in \{0,1\}}
\left(
1-\mathrm{GIoU}(\hat{\mathbf{B}}_i,\mathbf{B}_i)
\right),
\end{equation}

\begin{equation}
\mathcal{L}_{\mathrm{consistency}}
=
\frac{1}{2B}
\sum_{i \in \{0, 1\}}
\mathrm{L1}\Big(
\mathrm{normalize}(\hat{\mathbf{B}}_i), \mathrm{normalize}(\mathrm{bbox\_from\_mask}(\hat{\mathbf{M}}_i))
\Big).
\end{equation}

The total box loss is a weighted sum of these components,
\begin{align}
    \mathcal{L}_{\mathrm{box}} =& \lambda_\mathrm{IoU} \mathcal{L}_{\mathrm{IoU}} + \lambda_\mathrm{loc} \mathcal{L}_{\mathrm{loc}} + \lambda_\mathrm{size} \mathcal{L}_{\mathrm{size}} + \lambda_\mathrm{consistency} \mathcal{L}_{\mathrm{consistency}}.
\end{align}

The final mask loss combines point-sampled and region-aware components:
\begin{equation}
    \mathcal{L}_{\mathrm{mask}} = \alpha_\mathrm{point} \mathcal{L}_{\mathrm{point}} + \alpha_\mathrm{region} \mathcal{L}_{\mathrm{region}}.
\end{equation}

Finally, the total training loss is a weighted combination of mask and box supervision, reflecting their complementary roles in co-visible region learning:
\begin{equation}
\mathcal{L}_{\mathrm{total}}
=
\frac{\rho}{1+\rho}\mathcal{L}_{\mathrm{mask}}
+
\frac{1}{1+\rho}\mathcal{L}_{\mathrm{box}}.
\end{equation}

Point sampling emphasizes uncertain boundary regions, region weighting focuses supervision on in-box pixels, and mask--box consistency enforces dual-direction constraints, allowing mask predictions to refine box regression while boxes provide geometric guidance to mask learning.

\section{Experiments}
\label{sec:experiments}

We evaluate SAMatcher for correspondence estimation under cross-view scale imbalance, a data-level condition in which unequal effective resolution and spatial support reduce the comparability of the paired observations.
The experiments include implementation details, quantitative comparisons with state-of-the-art methods, qualitative analyses, and ablation studies of key components.
All evaluations use a consistent SAMatcher backbone and follow the same evaluation protocols to ensure fair comparison.

\subsection{Implementation Details}

SAMatcher is implemented in PyTorch and trained end-to-end.
Unless otherwise specified, all experiments share the same architectural configuration and training protocol.

\textbf{Encoder and Feature Extraction.}
We adopt the image encoder from SAM-HQ~\citep{ke2023segment} as backbone for both views.
The encoder is based on a Vision Transformer and produces dense feature maps at a spatial resolution of $H' \times W'$.
Following common practice, the encoder weights are frozen during training.
The output feature dimension is set to $C=256$.

\textbf{Symmetric Cross-View Interaction.}
The proposed cross-view feature interaction module follows a DiT-style token interaction paradigm.
Features from the source and target images are flattened into token sequences and treated as two correlated modalities.
The module is structured into 4 hierarchical stages: the first 2 stages consist of DoubleBlocks for cross-view interaction, followed by two stages of SingleBlocks for view-specific refinement.
The depths of the DoubleBlock stages are set to $[4,4]$, and the SingleBlock stages are set to $[4,2]$, resulting in a total of 14 transformer blocks along each view-specific path.
Correspondingly, the number of attention heads is scaled progressively across the four stages as 
$[8,16,32,64]$, allowing the model to focus on fine-grained correlations in early layers and global semantic coherence in deeper layers.
Each block adopts window-based multi-head self-attention with a window size of $8$.
To ensure training stability during cross-view interaction, we utilize cosine attention in DoubleBlocks, while SingleBlocks employ standard Flash Attention for efficiency.

\textbf{View-Aware Positional Encoding.}
View-aware rotary positional encoding is applied to the query and key projections in all attention layers within the interaction module.
A small view-dependent offset $\delta_v$ is added to the rotation angles to distinguish tokens from different views while preserving relative spatial encoding.
We set $\delta_v=0.1$ and share the same offset across all layers.

\textbf{Prompt and Decoder Design.}
Co-visible masks are predicted using the original SAM mask decoder, conditioned on fused features and shared learnable sparse and dense prompts.
To estimate spatial extents, we introduce a dedicated box decoder that operates exclusively on learnable box tokens.
Specifically, one box token is initialized for each view and updated through cross-attention, and its final embedding is decoded into a bounding box.
This design decouples geometric localization from dense feature decoding while enabling joint optimization through shared supervision.

\textbf{Training Setup.}
The model is trained using the joint loss described in Sec.~\ref{sec:losses}.
We set the loss weights to $\alpha_\mathrm{point}=1.0$, $\alpha_\mathrm{region}=1.5$, and the mask-box balance ratio $\rho=1.0$.
Point-based mask supervision samples $12,544$ points $(112\times 112)$ per image based on prediction uncertainty.

Training is performed using the AdamW optimizer with a learning rate of $3 \times 10^{-4}$ and a weight decay of $0.1$.
The batch size is set to $12$, and the model is trained for $50$ epochs.

\subsection{Datasets} \label{sec:datasets}

\textbf{MegaDepth} \citep{MegaDepthLi18}.
We use MegaDepth for both training and quantitative evaluation, as it provides ground-truth correspondences along with sparse 3D reconstructions and depth maps from COLMAP \citep{schonberger2016structure,schoenberger2016mvs}.
The dataset contains approximately 1M images across 196 outdoor scenes, covering diverse lighting conditions, scales, and viewpoints.

Following the SCoDe protocol for evaluating cross-view scale imbalance~\citep{pan2025scale}, we select 10 scenes for testing and randomly sample 500 image pairs with sufficient co-visible regions. These pairs expose the matcher to unequal observation quality across views while retaining enough shared scene content for geometric evaluation.
All experiments are conducted at a resolution of $1024 \times 1024$ unless otherwise specified.
Ground-truth co-visible regions are computed via ray-casting using camera parameters and depth maps, while negative pairs are sampled from unrelated scenes.

\textbf{ScanNet} \citep{dai2017scannet}.
We use ScanNet for qualitative evaluation of generalization to indoor environments.
It contains 2.5M images from 1,513 RGB-D scans, featuring cluttered scenes and limited viewpoint overlap.

\textbf{GL3D} \citep{shen2018mirror,luo2018geodesc}.
GL3D is used for qualitative evaluation in large-scale outdoor settings.
It consists of high-resolution aerial images with significant viewpoint and scale variations, representative of drone-based photogrammetry scenarios.

Both ScanNet and GL3D lack ground-truth co-visible region annotations and are therefore used only for qualitative analysis.

\subsection{Evaluation Metrics}

We evaluate SAMatcher on co-visible region detection and correspondence estimation under severe scale imbalance, with metrics reflecting both detection quality and geometric consistency. 
Intersection-over-Union (IoU) between predicted and ground-truth masks or bounding boxes is used primarily in ablation studies to assess region-level accuracy. 
For correspondence evaluation, we follow standard practice~\citep{sun2021loftr,sarlin2020superglue} and recover camera poses to compute rotation and translation errors. 
From these, we report the Area Under the Curve (AUC), Accuracy (Acc), and mean Average Accuracy (mAA), which summarize the recall computed at pixel error thresholds of 5, 10, and 20, providing a quantitative assessment of matching robustness under increasingly strict error criteria.
In addition, Matching Score (MS) denotes the mean confidence of all inlier correspondences, and Precision (P) measures the fraction of correspondences whose epipolar error falls below a threshold after pose recovery, quantifying the reliability and geometric consistency of predicted matches. 
Together, these metrics offer a comprehensive evaluation of SAMatcher's ability to detect co-visible regions and establish robust, geometrically consistent correspondences.

{
\renewcommand{\multirowsetup}{\centering}
\begin{table}[H]
\centering
\caption{
\textbf{Evaluation on MegaDepth under cross-view scale imbalance.} Each row corresponds to a complete matching pipeline. Specifically, we consider combinations of feature extractors and matchers, including SIFT~\citep{lowe2004distinctive}, SuperPoint (SP)~\citep{detone2018superpoint}, DISK~\citep{tyszkiewicz2020disk}, D2-Net (D2)~\citep{dusmanuD2netTrainableCNN2019}, ContextDesc (CON)~\citep{luo2019contextdesc}, R2D2~\citep{revaud2019r2d2},
LoFTR~\citep{sun2021loftr}, and RoMa~\citep{edstedt2024roma}. These are paired with either Nearest Neighbor (NN) or SuperGlue (SG)~\citep{sarlin2020superglue} for matching, except LoFTR and RoMa, which are end-to-end dense matching frameworks.
}
\label{tab:eval}
\resizebox{\linewidth}{!}{
\begin{tabular}{cl ccc cccc ccc c c}
    \toprule
    \multicolumn{2}{c}{\multirow{2}{*}{\textbf{Methods}}} & \multicolumn{3}{c}{\textbf{AUC}} & \multicolumn{4}{c}{\textbf{Acc}} & \multicolumn{3}{c}{\textbf{mAA}} & \multirow{2}{*}{\textbf{P}} & \multirow{2}{*}{\textbf{MS}} \\ 
    \cmidrule(l){3-12}
    \multicolumn{2}{c}{} & \multicolumn{1}{l}{@5} & @10 & @20 & @5 & @10 & @15 & @20 & @5 & @10 & @20 &  &  \\ 
    \midrule
    \multirow{3}{*}{\textbf{CON+NN}} & \textbf{-} & 28.07 & 43.91 & 58.91 & 49.80 & 66.60 & 74.80 & 77.80 & 49.80 & 58.20 & 67.25 & 70.76 & 9.77 \\
     & \textbf{+OETR} & 30.74 & 47.70 & 63.45 & 52.00 & 72.40 & 79.80 & 84.20 & 52.00 & 62.20 & 72.10 & 75.66 & 14.27 \\
     & \textbf{+SAMatcher} & \textbf{32.97} & \textbf{50.46} & \textbf{65.79} & \textbf{56.80} & \textbf{73.40} & \textbf{81.20} & \textbf{87.00} & \textbf{56.80} & \textbf{65.10} & \textbf{74.60} & \textbf{79.84} & \textbf{91.31} \\ 
    \midrule
    \multirow{3}{*}{\textbf{D2+NN}} & \textbf{-} & 0.85 & 1.40 & 2.29 & 1.27 & 2.29 & 3.31 & 3.82 & 1.27 & 1.78 & 2.67 & 9.34 & 2.23 \\
     & \textbf{+OETR} & 7.33 & 12.04 & 18.11 & 13.32 & 20.34 & 23.97 & 27.85 & 13.32 & 16.83 & 21.37 & 34.18 & 3.07 \\
     & \textbf{+SAMatcher} & \textbf{11.16} & \textbf{19.02} & \textbf{28.37} & \textbf{20.93} & \textbf{31.40} & \textbf{37.91} & \textbf{42.56} & \textbf{20.93} & \textbf{26.16} & \textbf{33.20} & \textbf{49.82} & \textbf{89.99} \\ 
    \midrule
    \multirow{3}{*}{\textbf{DISK+NN}} & \textbf{-} & 5.24 & 7.92 & 10.36 & 8.75 & 11.41 & 12.47 & 14.32 & 8.75 & 10.08 & 11.74 & 18.77 & 0.82 \\
     & \textbf{+OETR} & 20.63 & 30.06 & 38.10 & 34.58 & 43.03 & 45.52 & 47.76 & 34.58 & 38.81 & 42.72 & 56.70 & 4.88 \\
     & \textbf{+SAMatcher} & \textbf{27.78} & \textbf{40.71} & \textbf{53.36} & \textbf{45.52} & \textbf{59.34} & \textbf{66.75} & \textbf{69.57} & \textbf{45.52} & \textbf{52.43} & \textbf{60.29} & \textbf{75.09} & \textbf{89.65} \\ 
    \midrule
    \multirow{3}{*}{\textbf{DISK+SG}} & \textbf{-} & 25.29 & 37.49 & 50.06 & 43.37 & 55.82 & 62.45 & 68.27 & 43.37 & 49.60 & 57.48 & 39.50 & 9.62 \\
     & \textbf{+OETR} & 29.66 & 44.45 & 58.74 & 50.80 & 66.40 & 73.00 & 78.20 & 50.80 & 58.60 & 67.10 & 69.48 & 15.81 \\
     & \textbf{+SAMatcher} & \textbf{32.27} & \textbf{48.14} & \textbf{62.79} & \textbf{54.60} & \textbf{70.80} & \textbf{77.20} & \textbf{82.80} & \textbf{54.60} & \textbf{62.70} & \textbf{71.35} & \textbf{78.39} & \textbf{62.82} \\ 
    \midrule
    \multirow{3}{*}{\textbf{SIFT+NN}} & \textbf{-} & 5.85 & 11.13 & 19.79 & 11.20 & 20.60 & 28.80 & 35.00 & 11.20 & 15.90 & 23.90 & 17.06 & 85.93 \\
     & \textbf{+OETR} & 10.05 & 16.91 & 26.47 & 18.20 & 28.80 & 36.60 & 41.40 & 18.20 & 23.50 & 31.25 & 23.32 & 85.40 \\
     & \textbf{+SAMatcher} & \textbf{10.64} & \textbf{20.64} & \textbf{34.02} & \textbf{22.60} & \textbf{37.80} & \textbf{47.20} & \textbf{55.20} & \textbf{22.60} & \textbf{30.20} & \textbf{40.70} & \textbf{27.62} & \textbf{93.61} \\ 
    \midrule
    \multirow{3}{*}{\textbf{R2D2+NN}} & \textbf{-} & 20.47 & 33.45 & 47.27 & 37.60 & 52.60 & 61.60 & 68.00 & 37.60 & 45.10 & 54.95 & 52.89 & 87.81 \\
     & \textbf{+OETR} & 30.34 & 46.03 & 62.22 & 51.40 & 70.00 & 79.20 & 83.60 & 51.40 & 60.70 & 71.05 & 73.59 & 91.95 \\
     & \textbf{+SAMatcher} & \textbf{34.12} & \textbf{52.54} & \textbf{69.07} & \textbf{59.80} & \textbf{78.60} & \textbf{86.00} & \textbf{90.00} & \textbf{59.80} & \textbf{69.20} & \textbf{78.60} & \textbf{82.59} & \textbf{97.23} \\ 
    \midrule
    \multirow{3}{*}{\textbf{SP+NN}} & \textbf{-} & 3.74 & 5.81 & 9.64 & 5.60 & 10.40 & 13.40 & 17.00 & 5.60 & 8.00 & 11.60 & 17.31 & 3.94 \\
     & \textbf{+OETR} & 13.43 & 21.74 & 31.63 & 24.60 & 35.60 & 41.60 & 45.80 & 24.60 & 30.10 & 36.90 & 40.11 & 9.54 \\
     & \textbf{+SAMatcher} & \textbf{22.53} & \textbf{36.18} & \textbf{50.42} & \textbf{40.80} & \textbf{56.60} & \textbf{65.60} & \textbf{70.20} & \textbf{40.80} & \textbf{48.70} & \textbf{58.30} & \textbf{60.62} & \textbf{90.61} \\ 
    \midrule
    \multirow{3}{*}{\textbf{LoFTR}} & \textbf{-} & 34.36 & 48.33 & 60.79 & 53.80 & 68.20 & 73.20 & 77.20 & 53.80 & 61.00 & 68.10 & 55.61 & 43.91 \\
     & \textbf{+OETR} & 36.64 & 53.04 & 67.30 & 61.00 & 75.40 & 81.60 & 85.80 & 61.00 & 68.20 & 75.95 & 74.12 & 50.06 \\
     & \textbf{+SAMatcher} & \textbf{42.34} & \textbf{58.93} & \textbf{73.74} & \textbf{66.20} & \textbf{83.40} & \textbf{89.00} & \textbf{92.60} & \textbf{66.20} & \textbf{74.80} & \textbf{82.80} & \textbf{93.34} & \textbf{55.11} \\ 
    \midrule
    \multirow{3}{*}{\textbf{SP+SG}} & \textbf{-} & 33.46 & 49.64 & 64.75 & 55.20 & 71.60 & 80.80 & 85.20 & 55.20 & 63.40 & 73.20 & 85.19 & 13.54 \\
     & \textbf{+OETR} & 35.69 & 51.89 & 66.36 & 58.00 & 74.80 & 80.60 & 86.20 & 58.00 & 66.40 & 74.90 & 90.41 & 22.66 \\
     & \textbf{+SAMatcher} & \textbf{36.41} & \textbf{54.43} & \textbf{69.97} & \textbf{62.60} & \textbf{79.40} & \textbf{85.20} & \textbf{90.40} & \textbf{62.60} & \textbf{71.00} & \textbf{79.40} & \textbf{94.96} & \textbf{79.49} \\
    \midrule
    \multirow{3}{*}{\textbf{RoMa}} & \textbf{-} & 50.22 & 64.25 & 77.76 & 72.40 & 81.48 & 85.46 & 91.44 & 72.40 & 76.94 & 82.69 & 96.03 & 93.50 \\
     & \textbf{+OETR} & 44.91 & 62.17 & 76.51 & 67.99 & 86.14 & 90.15 & \textbf{92.70} & 67.99 & 77.07 & 84.25 & 81.37 & 98.97 \\
     & \textbf{+SAMatcher} & \textbf{55.80} & \textbf{70.06} & \textbf{80.36} & \textbf{77.37} & \textbf{87.42} & \textbf{90.82} & 91.47 & \textbf{77.37} & \textbf{82.39} & \textbf{86.75} & \textbf{97.63} & \textbf{99.57} \\
    \bottomrule
\end{tabular}
}
\end{table}
}

\subsection{Evaluation on Image Matching under Scale Imbalance}

We evaluate SAMatcher on the MegaDepth benchmark under cross-view scale imbalance, where unequal effective resolution, information availability, and spatial support create an imbalance in data quality across the paired views. The resulting cross-view observations are difficult to match even when each image is individually clear. This setting closely reflects wide-baseline and cross-resolution matching in remote sensing and large-scale scene reconstruction.
Following standard practice, we report AUC, accuracy (Acc), and mean Average Accuracy (mAA) at multiple thresholds.
Table~\ref{tab:eval} compares SAMatcher with baseline pipelines and OETR-enhanced variants across a broad spectrum of feature extractors and matching strategies.
Here, OETR~\citep{chen2022guide} denotes a region-level matching prior that constrains correspondences using predicted object-level bounding regions, serving as a coarse co-visibility baseline.

For pipelines based on nearest neighbor matching, SAMatcher consistently yields substantial and stable improvements.
Notably, descriptors that are particularly sensitive to scale variation and background clutter, such as D2-Net and SuperPoint, benefit the most.
For D2-Net+NN, AUC@20 increases from 2.29 to 28.37 and mAA@20 from 2.67 to 33.20, indicating a qualitative shift from near-failure to reliable matching.
Similar trends are observed for DISK+NN and SP+NN, where SAMatcher improves both low-threshold accuracy and overall mAA by large margins.
These results suggest that NN-based matching is especially vulnerable when scale imbalance makes the input views unequal in usable image information. Correspondences formed in non-overlapping or weakly supported regions are effectively suppressed by explicit co-visible region modeling.

When combined with learning-based matching frameworks, SAMatcher continues to provide consistent gains, although the relative improvement is more moderate.
For DISK+SG, SAMatcher improves AUC@20 from 50.06 to 62.79 and mAA@20 from 57.48 to 71.35, while for SP+SG, mAA@20 increases from 73.20 to 79.40.
These improvements indicate that even when relational reasoning and contextual aggregation are already present in the matcher, explicitly enforcing cross-view co-visibility helps it handle the data-quality imbalance between views by introducing complementary geometric and semantic constraints.
Rather than replacing learned correspondence reasoning, SAMatcher refines the candidate space by filtering out structurally inconsistent regions before matching, leading to more reliable pose estimation.

The benefit of SAMatcher is further evident when applied to dense end-to-end matching. For LoFTR, which directly operates on dense feature correlations, SAMatcher improves AUC@20 from 60.79 to 73.74 and mAA@20 from 68.10 to 82.80. Similarly, when integrated with RoMa, SAMatcher increases AUC@20 from 77.76 to 80.36 and mAA@20 from 82.69 to 86.75. These consistent improvements indicate that dense matching alone does not fully resolve ambiguities caused by partial overlap and unequal observation quality under scale imbalance. By explicitly modeling co-visible regions as an intermediate representation, SAMatcher provides a higher-level structural prior for handling such imbalanced cross-view observations.

To assess generality beyond learned representations, we also evaluate SAMatcher with classical SIFT features and nearest-neighbor matching.
Despite SIFT being explicitly designed for scale invariance, SAMatcher still improves AUC@20 from 19.79 to 34.02 and mAA@20 from 23.90 to 40.70.
Importantly, these gains are achieved without altering the detector or descriptor, confirming that SAMatcher addresses input-level data imbalance rather than modifying descriptor design.
This result highlights that co-visible region estimation handles a data-level aspect of the matching problem that is largely orthogonal to descriptor design.

Although OETR improves most of the evaluated matching methods, it degrades RoMa, reducing AUC@5 from 50.22 to 44.91 and Acc@5 from 72.40 to 67.99.

One possible explanation is that RoMa already possesses implicit co-visibility reasoning, while OETR's box-level overlap constraints may not be fully compatible with its dense matching mechanism. In contrast, SAMatcher provides finer-grained and complementary co-visibility guidance through joint mask--box modeling.

Overall, the results in Table~\ref{tab:eval} demonstrate that SAMatcher consistently improves correspondence quality across diverse matching paradigms under scale imbalance.
The improvements show that explicit cross-view fusion and co-visibility modeling can mitigate unequal observation resolution, spatial support, and partial overlap between paired views. Such cross-view data-quality imbalance is common in aerial and satellite imagery.
These findings support explicit co-visibility modeling as a principled and broadly applicable strategy for robust image matching in large-scale and remote sensing scenarios.

\begin{figure}[htbp]
    \centering
    \includegraphics[width=0.9\linewidth]{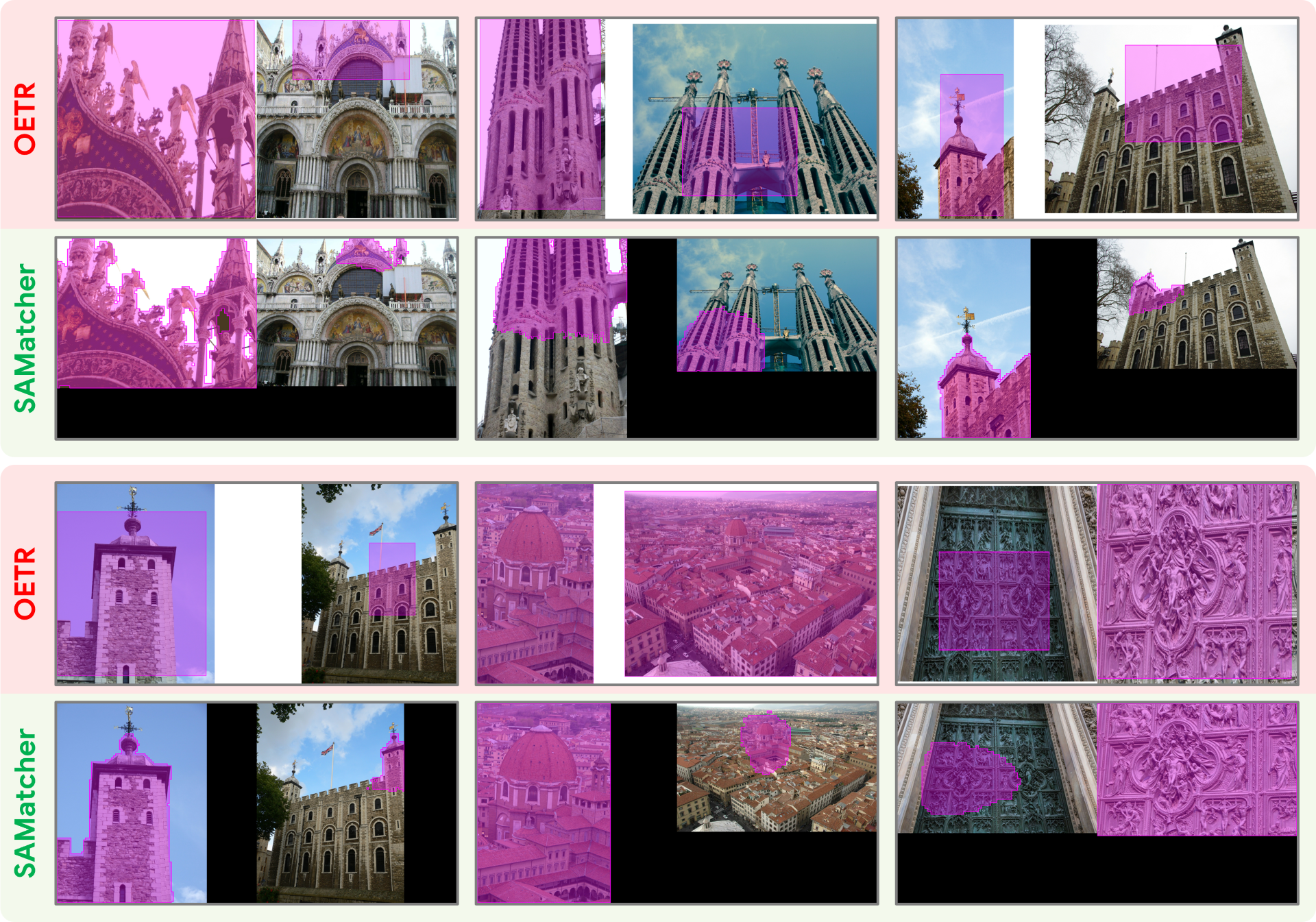}
    \caption{
        \textbf{Qualitative comparison of co-visible region detection.}
        For each image pair, we show OETR
        \citep{chen2022guide} box-only predictions and SAMatcher
        mask predictions, with masks overlaid as semi-transparent
        purple regions. While OETR provides coarse bounding boxes,
        SAMatcher produces accurate and consistent co-visible regions
        across views, even under severe scale imbalance, viewpoint changes, and partial overlap.
    }
    \label{fig:covis_mask}
\end{figure}

\subsection{Qualitative Analysis of Co-Visible Region Modeling}

\subsubsection{Co-Visible Region Detection}

Fig.~\ref{fig:covis_mask} presents qualitative results of co-visible region detection on representative MegaDepth image pairs with cross-view scale imbalance.
Although corresponding structures occupy substantially different spatial extents across views, SAMatcher consistently highlights mutually observable regions while suppressing view-specific background and occlusions.
The predicted masks do not simply follow local saliency or semantic prominence; they remain consistent despite the imbalance in observation quality caused by unequal effective resolution and spatial support, activating regions only when the same content remains usable in both views.

This behavior indicates that co-visibility is learned as a relational property between views rather than an intrinsic feature of individual images.
Such relational modeling is critical under severe scale imbalance, where the views contain unequal image information and appearance similarity alone cannot distinguish valid correspondences from coincidental matches.
By explicitly identifying co-visible regions prior to correspondence estimation, SAMatcher establishes a structured intermediate representation that reduces the ambiguity of subsequent matching.
These observations provide direct evidence that the symmetric cross-view interaction successfully embeds co-visibility cues into the feature representation, laying a reliable foundation for downstream correspondence estimation.

\subsubsection{Region-Guided Correspondence Filtering}

\begin{figure}
\centering
\includegraphics[width=\linewidth]{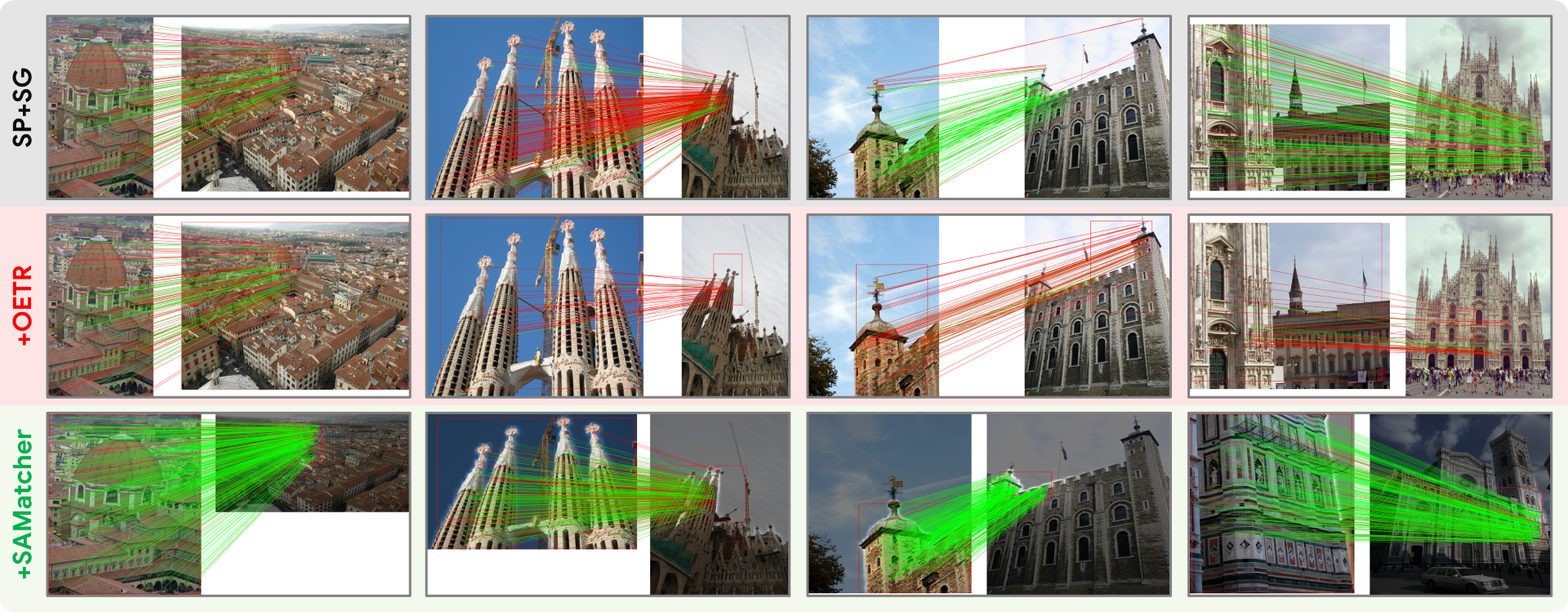}
\caption{
    \textbf{Region-guided correspondence comparison.}
    SP+SG, +OETR, and +SAMatcher.
    Green lines denote correct matches, red lines incorrect ones.
    Under severe scale imbalance, OETR often predicts inaccurate or missing regions, while SAMatcher identifies valid co-visible regions and yields more reliable correspondences.
}
\label{fig:covis_match}
\end{figure}

Fig.~\ref{fig:covis_match} visualizes correspondence results guided by predicted co-visible regions.
Compared to unconstrained matching, valid correspondences are densely concentrated within the predicted co-visible areas, while erroneous matches are largely suppressed outside these regions.
False correspondences mainly arise in areas lacking mutual visibility, such as background clutter, repeated textures, or view-specific structures.

Under cross-view scale imbalance, OETR relies on coarse bounding boxes that often fail to capture valid regions and may introduce incorrect spatial priors.
SAMatcher instead fuses cross-view features to provide more accurate co-visible region estimates, enabling effective filtering of invalid matches when corresponding content occupies substantially different spatial extents.

Constraining correspondence search to jointly supported regions effectively reduces the combinatorial space of potential matches.
This region-guided filtering alleviates common failure modes under large viewpoint or scale changes, where local descriptors may appear similar despite lacking geometric consistency.
The observed reduction in spurious matches corresponds to consistent improvements in AUC, accuracy, and mAA reported in Table~\ref{tab:eval}, demonstrating that explicit co-visible region modeling directly enhances correspondence reliability rather than merely refining intermediate features.

\subsubsection{Complementarity of Mask and Box Supervision}

\begin{figure}
\centering
\includegraphics[width=\linewidth]{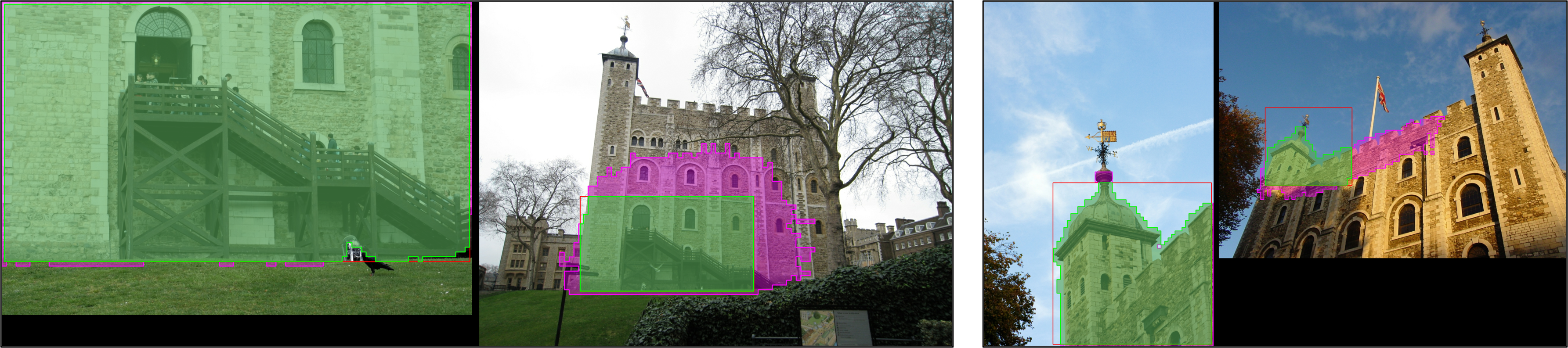}
\caption{
    \textbf{Complementarity of mask and box predictions.}
    Masks (magenta) provide high recall but coarse coverage, while boxes (red) offer precise localization.
    Constraining masks with boxes yields refined co-visible regions (green), improving correspondence reliability.
}
\label{fig:mask_box_comp}
\end{figure}

Fig.~\ref{fig:mask_box_comp} illustrates the complementary roles of mask and box predictions in co-visible region modeling.
The predicted masks generally favor high recall by covering most potentially co-visible areas, whereas the corresponding bounding boxes provide compact geometric localization.
Although either representation alone is insufficient, their combination effectively balances coverage and precision.

Specifically, intersecting the predicted masks with the corresponding bounding boxes suppresses over-extended regions while preserving valid co-visible content, resulting in more reliable spatial support for correspondence estimation.
This complementary behavior justifies the proposed joint supervision strategy, where dense mask prediction and object-level localization mutually regularize each other.
Consequently, SAMatcher produces more accurate co-visible regions under cross-view scale imbalance and provides stronger spatial priors for downstream matching.

\subsection{Generalization}

We evaluate the zero-shot generalization of SAMatcher on two previously unseen datasets with markedly different scene characteristics. ScanNet~\citep{dai2017scannet} consists of indoor environments, whereas GL3D~\citep{shen2018mirror,luo2018geodesc} primarily contains outdoor scenes, including aerial imagery. No fine-tuning is performed on either dataset.

\begin{figure}
\centering
\includegraphics[width=\linewidth]{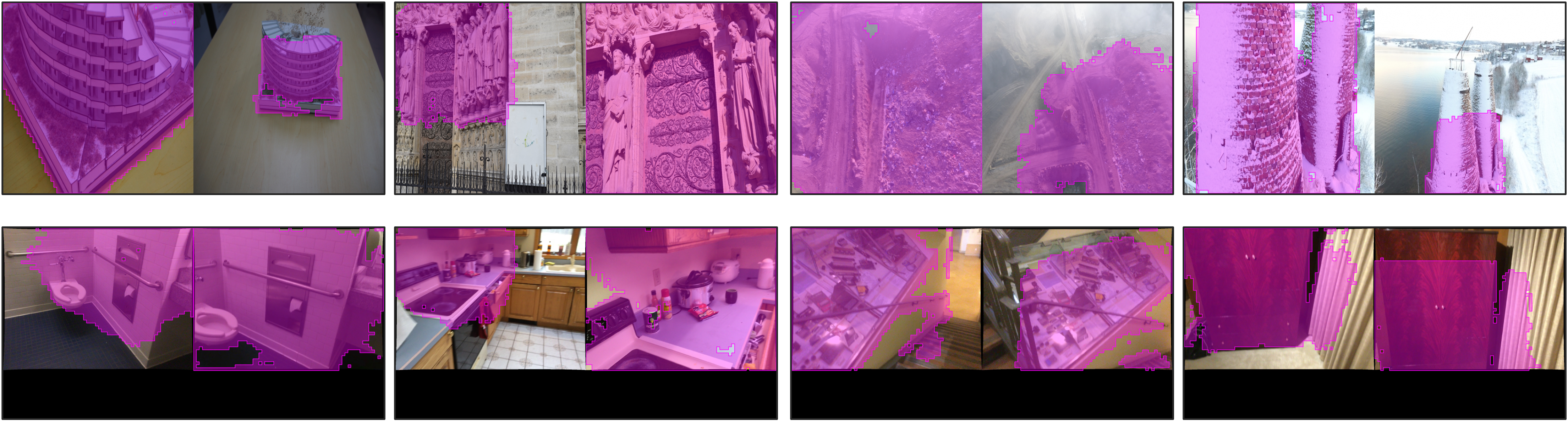}
\caption{
\textbf{Zero-shot generalization on unseen datasets.}
Top: GL3D (outdoor aerial scenes). Bottom: ScanNet (indoor environments).
Predicted co-visible regions are overlaid as semi-transparent magenta masks.
SAMatcher consistently captures mutually observable regions under cross-view scale imbalance while suppressing non-overlapping content across domain shifts.
}
\label{fig:generalization}
\end{figure}

As shown in Fig.~\ref{fig:generalization}, SAMatcher consistently identifies co-visible regions across image pairs with different inter-view scales while suppressing content outside the overlapping field of view, despite substantial domain shifts. On GL3D, it remains robust when scale imbalance produces large differences in the amount of usable image information across views. On ScanNet, it effectively suppresses occluded regions and background clutter, retaining regions that are geometrically consistent across the two observations.

These qualitative results indicate that SAMatcher captures co-visibility as a transferable cross-view relationship rather than relying primarily on dataset-specific appearance priors. This ability remains stable when scale imbalance creates unequal observation quality across paired views, supporting robust correspondence reasoning across diverse visual domains.

\subsection{Ablation Study}
\label{sec:ablation}

\begin{table}[h!]
    \centering
    \caption{Ablation results on MegaDepth.}
    \label{tab:ablation}
    \begin{tabular}{lcc}
        \toprule
        Method & Mask IoU (\%) & Box IoU (\%) \\
        \midrule
        w/o Symmetric Interaction & 75.06 & 71.90 \\
        w/o Joint Constraint Optimization & 76.67 & 80.68 \\
        w/o View Offset & 77.15 & 80.74 \\
        \midrule
        Full SAMatcher & \textbf{78.86} & \textbf{82.38} \\
        \bottomrule
    \end{tabular}%
\end{table}

We perform a controlled ablation to quantify the contribution of each core component in SAMatcher.
All variants share the same backbone, optimization, and loss weights.
Performance is measured via mask IoU and box IoU, as shown in Table~\ref{tab:ablation}.

We evaluate three variants: (1) without symmetric cross-view interaction, (2) without joint mask--box consistency, and (3) without view offset modeling.
In variant (2), mask and box branches are optimized independently.

Removing symmetric interaction reduces mask and box IoU by 3.8\% and 10.5\%, respectively, highlighting the importance of cross-view interaction for reasoning about imbalanced observations.
Disabling joint constraint optimization lowers mask IoU by 2.2\% and box IoU by 1.7\%, confirming that cross-branch consistency provides effective regularization.
Excluding view offset modeling decreases mask IoU by 1.7\% and box IoU by 1.6\%, showing that relative viewpoint encoding improves localization of shared regions.

The full SAMatcher reaches 78.86\% mask IoU and 82.38\% box IoU, validating that symmetric interaction, joint constraints, and view offset modeling collectively enable precise co-visible region prediction.

\section{Conclusion}
\label{sec:conclusion}

We present SAMatcher, a modular co-visibility modeling framework for feature matching under cross-view scale imbalance. Scale-imbalanced image pairs are low-quality data for matching because local correspondence evidence becomes less comparable across views. SAMatcher explicitly fuses features from the two views to infer dense regions whose visual evidence remains jointly reliable and converts them into view-specific cropping regions for downstream matching. Unlike end-to-end matchers that couple visibility reasoning with correspondence prediction, SAMatcher isolates co-visible region extraction as an independent and reusable module. Experiments following the SCoDe scale-imbalance evaluation protocol demonstrate consistent improvements across classical, learning-based, and dense matching pipelines. These results show that explicit cross-view fusion and co-visibility modeling improve point-level matching robustness under severe scale imbalance and provide a practical direction for handling quality-imbalanced multi-view data.
\vspace*{-1em}
\bibliography{main}


\end{document}